\documentclass{article}
\usepackage[utf8]{inputenc}
\usepackage{url}
\usepackage{floatrow}
\usepackage{caption}
\usepackage{hyperref}
\usepackage{graphicx}
\usepackage{verbatim}
\usepackage{xspace}
\usepackage{subcaption}
\usepackage[dvipsnames]{xcolor}
\usepackage[export]{adjustbox}

\usepackage{graphicx,calc}
\newlength\myheight
\newlength\mydepth
\settototalheight\myheight{Xygp}
\definecolor{CardinalRed}{rgb}{0.55,.08,0.08}
\definecolor{BurntOrange}{rgb}{0.75,.34,0.00}
\definecolor{DarkBlue}{rgb}{0,.37,0.53}

\hypersetup{
    colorlinks=true,
    linkcolor=DarkBlue,
    filecolor=magenta,      
    urlcolor=BurntOrange,
}

\usepackage{adjustbox}

\newcommand{\projectName}{\texttt{robosuite}\xspace}

\newcommand{\class}[1]{{\textbf{\textcolor{CardinalRed}{#1}}}}

\title{\projectName{}: A Modular Simulation Framework and Benchmark for Robot Learning}
\author{\small Yuke Zhu$^\clubsuit$ \quad Josiah Wong$^\clubsuit$ \quad Ajay Mandlekar$^\clubsuit$ \quad Roberto Mart\'{i}n-Mart\'{i}n$^\clubsuit$\footnote{$\clubsuit$: founding members who initiate and lead this project} \\ \small Abhishek Joshi$^\diamondsuit$ \quad Kevin Lin $^\diamondsuit$ \quad Abhiram Maddukuri$^\diamondsuit$ \\ \small Soroush Nasiriany$^\diamondsuit$ \quad \small Yifeng Zhu$^\diamondsuit$\footnote{$\diamondsuit$: core members who make significant contributions (in alphabetical order)}}
\date{\href{https://robosuite.ai}{\textbf{robosuite.ai}}}

\begin{document}

\renewcommand{\arraystretch}{1.3}

\maketitle

\begin{figure}
    \centering
    \includegraphics[width=1.0\linewidth]{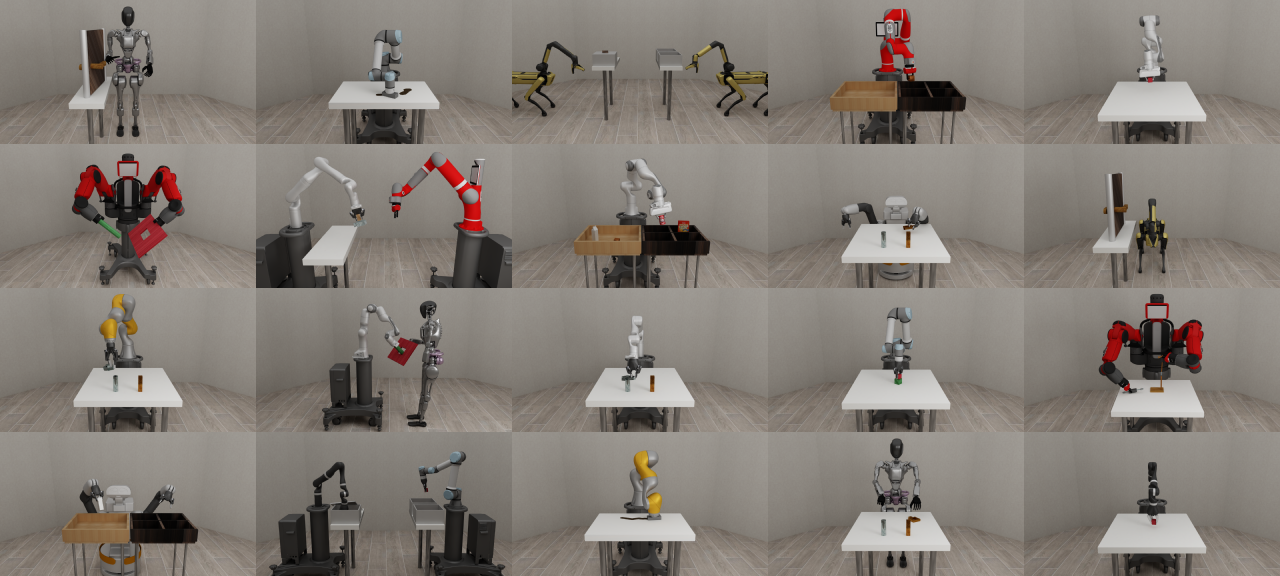}
    \caption{Procedurally generated robotic environments with \projectName{} APIs}
    \label{fig:my_label}
\end{figure}

\begin{abstract}
    \projectName{} is a simulation framework for robot learning powered by the MuJoCo physics engine. It offers a modular design for creating robotic tasks as well as a suite of benchmark environments for reproducible research. This paper discusses the key system modules and the benchmark environments of our new release \projectName{} v1.5. For the latest updates on \projectName{}, please visit our project website. 
\end{abstract}

\section{Introduction}

We introduce \projectName{}, a modular simulation framework and benchmark for robot learning. This framework is powered by the MuJoCo physics engine~\cite{todorov2012mujoco}, which performs fast physical simulation of contact dynamics. The overarching goal of this framework is to facilitate research and development of data-driven robotic algorithms and techniques. The development of this framework was initiated from the SURREAL project~\cite{corl2018surreal} on distributed reinforcement learning for robot manipulation, and is now part of the broader Advancing Robot Intelligence through Simulated Environments (ARISE) Initiative, with the aim of lowering the barriers of entry for cutting-edge research at the intersection of AI and Robotics.

Data-driven algorithms~\cite{kroemer2019review}, such as reinforcement learning~\cite{sutton2018reinforcement,kober2013reinforcement} and imitation learning~\cite{ravichandar2020recent}, provide a powerful and generic tool in robotics. These learning paradigms, fueled by new advances in deep learning, have achieved some exciting successes in a variety of robot control problems. Nonetheless, the challenges of reproducibility and the limited accessibility of robot hardware have impaired research progress~\cite{henderson2018deep}. In recent years, advances in physics-based simulations and graphics have led to a series of simulated platforms and toolkits~\cite{brockman2016openai,tassa2020dm_control,kolve2017ai2,dosovitskiy2017carla,xia2020interactive} that have accelerated scientific progress on robotics and embodied AI. Through the \projectName{} project we aim to provide researchers with:
\begin{enumerate}
    \item a modular design that offers great flexibility to create new robot simulation environments and tasks;
    \item a high-quality implementation of robot controllers and off-the-shelf learning algorithms to lower the barriers to entry;
    \item a standardized set of benchmark tasks for rigorous evaluation and algorithm development.
\end{enumerate}
Our new release of \projectName{} v1.5 contains ten robot models, nine gripper models, four base models, six body part controller modes, and nine standardized tasks. The body part controllers span the most frequently used action spaces (joint space and Cartesian space) and have been tested and used in previous projects~\cite{martin2019variable}. We use a composite controller interface to compose different body part controllers. \projectName{} also offers a modular design of APIs for building new environments with procedural generation. We highlight these primary features below:
\begin{enumerate}
    \item standardized tasks: a set of standardized manipulation tasks of large diversity and varying complexity and RL benchmarking results for reproducible research;
    \item procedural generation: modular APIs for programmatically creating new environments and new tasks as a combinations of robot models, arenas, and parameterized 3D objects;
    \item realistic composite robot controllers: a high-level composite controller that orchestrates a selection of body part controller implementations to command the robots in joint space and Cartesian space, in position, velocity or torque, including inverse kinematics control, and operational space control;
    \item multi-modal sensors: heterogeneous types of sensory signals, including low-level physical states, RGB cameras, depth maps, segmentation masks, and proprioception;
    \item human demonstrations: utilities for collecting human demonstrations (with keyboard, 3D mouse and GUI with cursor devices), replaying demonstration datasets, and leveraging demonstration data for learning.
\end{enumerate}
In the rest of the manuscript, we will describe the overall design of the simulation framework and the key system modules. We will also describe the benchmark tasks in the v1.0 \projectName{} release and benchmarking results of the most relevant state-of-the-art data-driven algorithms on these tasks.

\newcommand{\robotfigure}[2]{
\begin{figure}[H]
\floatbox[{\capbeside\thisfloatsetup{capbesideposition={right,top},
capbesidewidth=0.7\textwidth}}]{figure}[\FBwidth]
{\caption*{#1}}
{\adjincludegraphics[width=0.35\textwidth]{#2}}
\end{figure}
}

\section{System Modules}

\begin{figure}
    \centering
    \includegraphics[width=1.0\linewidth]{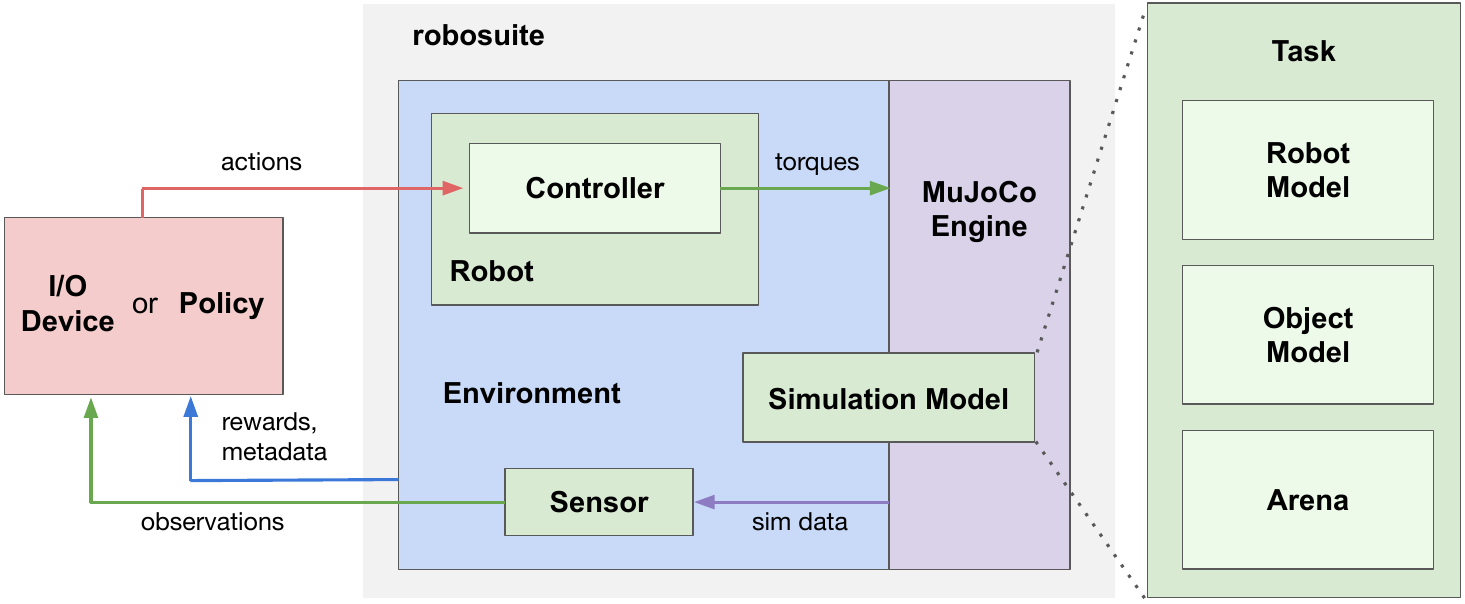}
    \caption{System diagram of \projectName{} modules. An actor (e.g. a Policy or a human using an I/O Device) generates actions commands and pass them to the \projectName{} Environment. The action is transformed by the controller of the robot into torque commands and executed by the MuJoCo physics engine. The result of the execution is observed via Sensors, that provide observations to the actors, together with reward and metadata from the Environment. The Environment is the result of instantiating a Task model composed of a Robot Model, an Arena, and, possibly, some objects defined by Object Models.}
    \label{fig:system_diagram}
\end{figure}

In this section we describe the overall system design of \projectName{}. \projectName{} offers two main APIs: 1) \textbf{Modeling APIs} to describe and define simulation environments and tasks in a modular and programmatic fashion, and 2) \textbf{Simulation APIs} to wrap around the physics engine and provide an interface for external actors (i.e. a \textbf{Policy} or an \textbf{I/O Device}) to execute actions, and receive observations and rewards. The Modeling APIs are used to generate a \textbf{Simulation Model} that can be instantiated by the MuJoCo engine~\cite{todorov2012mujoco} to create a simulation runtime, called \textbf{Environment}. The Environment generates observations through the \textbf{Sensors}, such as cameras and robot proprioception, and receives action commands from policies or I/O devices that are transformed from the original action space (e.g. joint velocities, or Cartesian positions) into torque commands by the \textbf{Controllers} of the \textbf{Robots}. The diagram above illustrates the key components in our framework and their relationships.

A simulation model is defined by a \class{Task} object, which encapsulates three essential constituents of robotic simulation: \class{RobotModel}(s), Object Model(s), and an \class{Arena}. A task may contain one or more robots, zero to many objects, and a single arena. The \class{RobotModel} object loads models of robots and their corresponding \class{GripperModel}(s) and \class{RobotBaseModel} from provided XML files. The Object Model defined by \class{MujocoObject} can be either loaded from 3D object assets or procedurally generated with programmatic APIs. The \class{Arena} object defines the workspace of the robot, including the environment fixtures, such as a tabletop, and their placements. The \class{Task} class recombines these constituents into a single XML object in MuJoCo’s \href{http://www.mujoco.org/book/XMLreference.html}{\textbf{MJCF modeling language}}. The generated MJCF object is passed to the MuJoCo engine through the \href{https://github.com/google-deepmind/mujoco}{\textbf{mujoco}} library that instantiates and initializes the simulation. The result of this instantiation is a MuJoCo runtime simulation object (the \class{MjSim} object) that contains the state of the simulator, and that will be interfaced via our Simulation APIs.

The \class{Environment} object (see Section~\ref{sec:environments}) provides \href{https://gym.openai.com/}{\textbf{OpenAI Gym}}-style APIs for external inputs to interface with the simulation. External inputs correspond to the action commands used to control the \class{Robot}(s) their grippers and their bases (see Section~\ref{sec:robots}), where the action spaces are specific to the \class{Controller}(s) used by the robots (see Section~\ref{sec:controllers}). For instance, for a joint-space position controller for a robot arm, the action space corresponds to desired position of each joint of the robot (dimensionality$=$number of degrees of freedom of the robot), and for an operational space controller, the action space corresponds to either desired 3D Cartesian position or desired full 6D Cartesian pose for the end-effector. These action commands can either be automatically generated by an algorithm (such as a deep neural network policy) or come from an I/O \class{Device}(s) for human teleoperation, such as the keyboard or a 3D mouse (see Section~\ref{sec:devices}). The controller of the robot is responsible for interpreting these action commands and transforming them into the low-level torques passing to the underlying physics engine (MuJoCo), which performs internal computations to determine the next state of the simulation. \class{Sensor}(s) retrieve information from the \class{MjSim} object and generate observations that correspond to the physical signals that the robots receive as response to their actions (see Section~\ref{sec:sensors}). Our framework supports multiple sensing modalities, such as RGB-D cameras, force-torque measurements, and proprioceptive data, allowing multimodal solutions to be developed. In addition to these sensory data, environments also provide additional information about the task progress and success conditions, including reward (for reinforcement learning) and other meta-data (e.g. task completion). In the following, we describe in detail the individual components of \projectName{}.

\subsection{Environments}
\label{sec:environments}

\class{Environments} provide the main APIs for external/user code to interact with the simulator and perform tasks. Each environment corresponds to a robotic task and provides a standard interface for an agent to interact with the environment.

\class{Environment}(s) are created by calling \texttt{make} with the name of the task (see Section~\ref{sec:task_desc} for a suite of standardized tasks provided in \projectName{}) and with a set of arguments that configure environment properties such as whether on-screen (\texttt{has\_renderer}) or off-screen rendering (\texttt{has\_offscreen\_renderer}) is used, whether the observation space includes physical states (\texttt{use\_object\_obs}) or image (\texttt{use\_camera\_obs}) observations, the control frequency (\texttt{control\_freq}), the episode horizon (\texttt{horizon}), and whether to shape rewards or use a sparse one (\texttt{reward\_shaping}). 

Environments have a modular structure, making it easy to construct new ones with different robot arms (\texttt{robots}), grippers (\texttt{gripper\_types}), and controllers (\texttt{controller\_configs}). Every environment owns its own MJCF model that sets up the MuJoCo physics simulation by loading the robots, the workspace, and the objects into the simulator appropriately. This MuJoCo simulation model is programmatically instantiated in the \texttt{\_load\_model} function of each environment, by creating an instance of the \class{Task} class.

Each \class{Task} class instance owns an \class{Arena} model, a list of \class{Robot} model instances, and a list of \class{Object} model instances. These are \projectName{} classes that introduce a useful abstraction in order to make designing scenes in MuJoCo easy. Every \class{Arena} is based off of an XML that defines the workspace (for example, table or bins) and camera locations. Every \class{Robot} is a MuJoCo model of each type of robot arm (e.g., Sawyer, Panda, etc.). Every \class{Object} model corresponds to a physical object loaded into the simulation (e.g., cube, pot with handles, etc.).

Each \class{Task} class instance also takes a \texttt{placement\_initializer} as input. The \texttt{placement\_initializer} determines the start state distribution for the environment by sampling a set of valid, non-colliding placements for all of the objects in the scene at the start of each episode (e.g., when \texttt{env.reset()} is called).

\subsection{Robots}
\label{sec:robots}

\begin{figure}
    \centering
    \includegraphics[width=\linewidth]{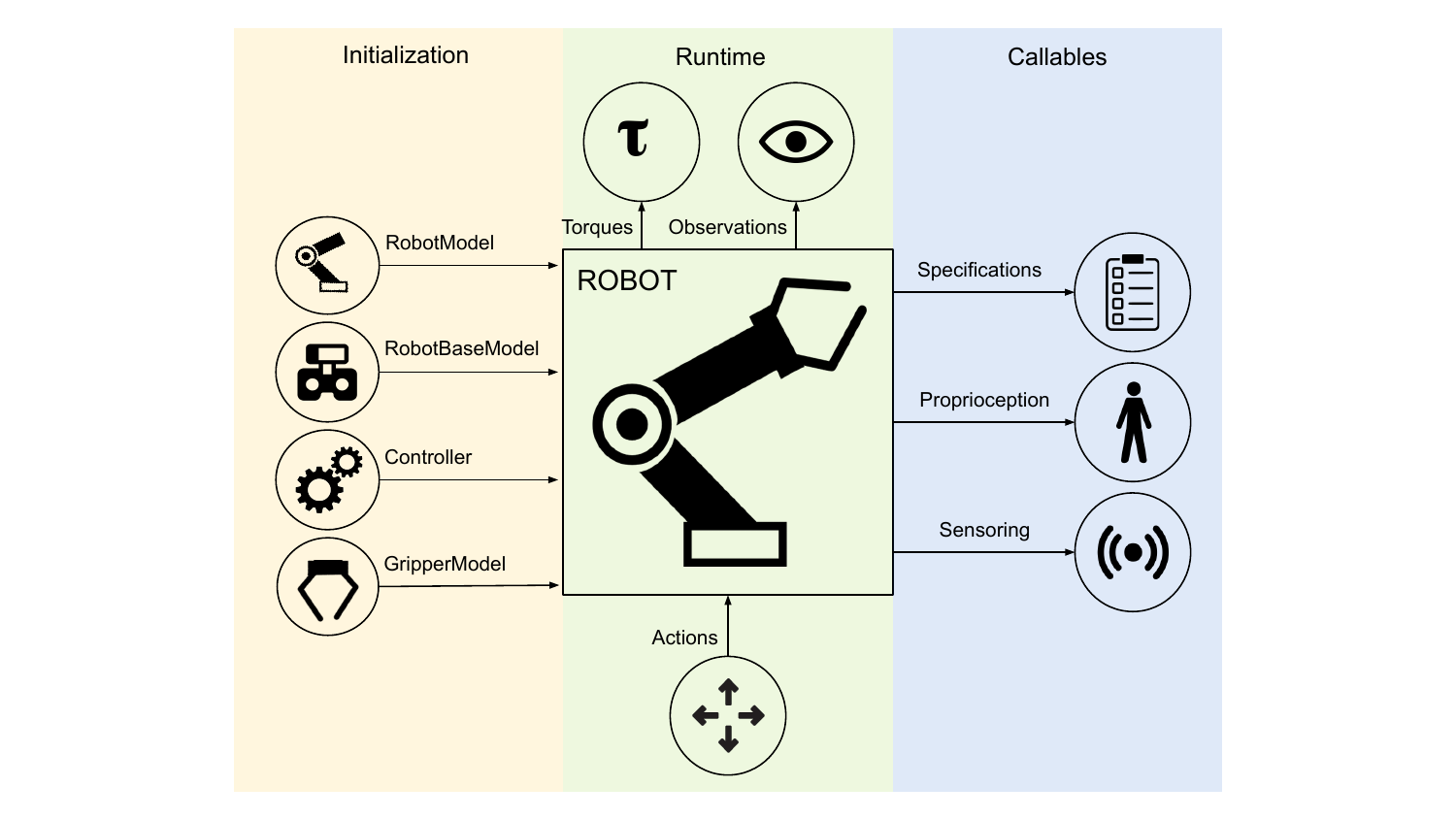}
    \caption{Overview of the Robot module's structure and usage. A \class{Robot} is initialized with appropriate models and controller, interacts with the environment during runtime, and can be accessed to retrieve relevant state information at any point during the simulation.}
    \label{fig:robot_module}
\end{figure}

\textbf{Robots} are a key component in \projectName{}, serving as the embodiment of the agent that interacts within the environment. \projectName{} captures this level of abstraction with the \class{Robot}-based classes, with support for both single-armed and bimanual variations, as well as robots with mobile manipulation capabilities, including both legged and wheeled variants. In turn, the Robot class is defined by a \class{RobotModel}, \class{GripperModel}(s) (with no gripper being represented by a dummy class), \class{RobotBaseModel}, and \class{Controller}(s). The high-level features of \projectName{}’s robots are described as follows:

\begin{itemize}
    \item \textbf{Diverse and Realistic Models:} the current version of \projectName{} provides models for 10 commercially-available robots (including the \textbf{humanoid} GR1 Robot), 9 grippers (including the Inspire \textbf{dexterous hand} model), 4 bases (including the Omron wheeled mobile base), and 6 body-part controllers, with model properties either taken directly from official product documentation or raw spec sheets. We also provide an extension package from the \href{https://github.com/ARISE-Initiative/robosuite_models}{\texttt{robosuite-models}\xspace} repository which currently includes additional 8 robots, 8 grippers, and 3 bases. This extension package must be installed separately and it is actively maintained. 
    \item \textbf{Modularized Support:} Robots are designed to be plug-and-play---any combinations of robots, grippers, bases, and controllers can be used, assuming the given environment is intended for the desired robot configuration. Because each robot is assigned a unique ID number, multiple instances of identical robots can be instantiated within the simulation without error.
    \item \textbf{Self-Enclosed Abstraction:} For a given task and environment, any information relevant to the specific robot instance can be found within the properties and methods within that instance. This means that each robot is responsible for directly setting its initial state within the simulation at the start of each episode, and also directly controls the robot in simulation via torques outputted by its controller’s transformed actions.
\end{itemize}

\projectName{} currently supports 10 commercially-available robot models. We briefly describe each individual model along with its features below:

\robotfigure{\textbf{\href{https://www.franka.de/technology}{Panda}} is a 7-DoF and relatively new robot model produced by Franka Emika, and boasts high positional accuracy and repeatability. A common choice for both simulated and real-robot research, we provide a substantial set of \hyperref[sec:benchmarking_experiments]{benchmarking experiments} using this robot. The default gripper for this robot is the \class{PandaGripper}, a parallel-jaw gripper equipped with two small finger pads, that comes shipped with the robot arm.}{robot_model_panda_v15.png}

\robotfigure{\textbf{\href{https://www.rethinkrobotics.com/sawyer}{Sawyer}} is Rethink Robotic's 7-DoF single-arm robot, which also features an additional 8th joint (inactive and disabled by default in \projectName{}) for swiveling its display monitor. Along with Panda, Sawyer serves as the second testing robot for our set of \hyperref[sec:benchmarking_experiments]{benchmarking experiments}. Sawyer's default \class{RethinkGripper} model is a parallel-jaw gripper with long fingers and useful for grasping a variety of objects.}{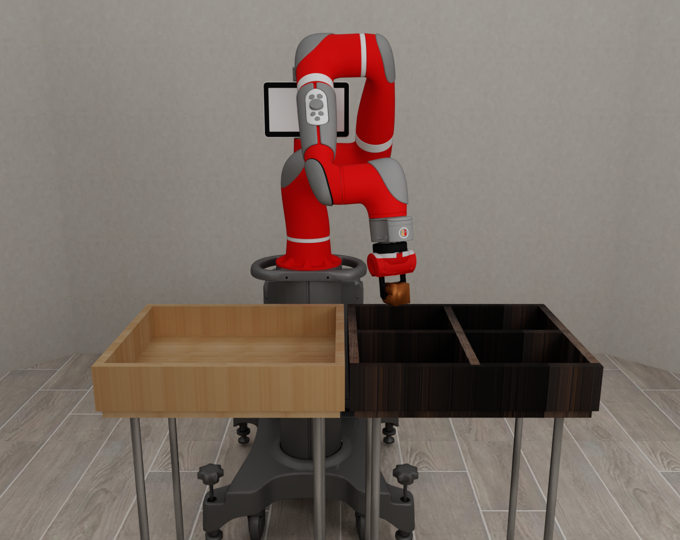}

\robotfigure{\textbf{\href{https://www.kuka.com/en-us/products/robotics-systems/industrial-lbr-iiwa}{IIWA}} is one of KUKA's industrial-grade 7-DoF robots, and is equipped with the strongest actuators of the group, with its per-joint torque limits exceeding nearly all the other models in \projectName{} by over twofold! By default, IIWA is equipped with the \class{Robotiq140Gripper}, \href{https://robotiq.com/products/2f85-140-adaptive-robot-gripper}{\textbf{Robotiq's 140mm variation}} of their multi-purpose two finger gripper models.}{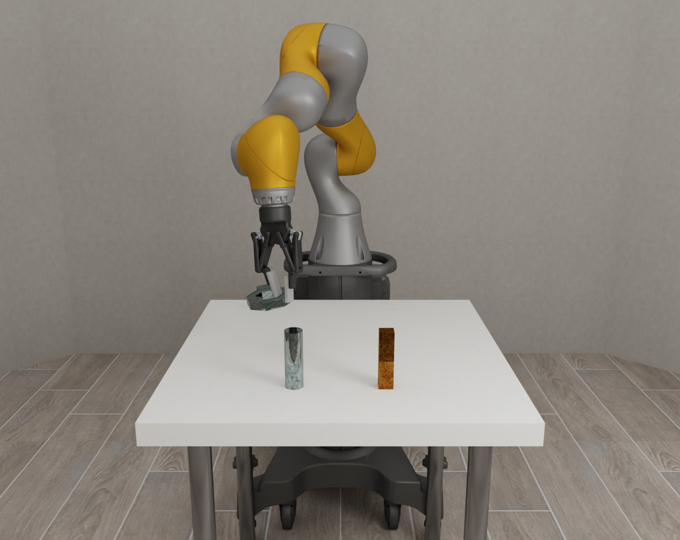}

\robotfigure{\textbf{\href{https://www.kinovarobotics.com/en/products/assistive-technologies/kinova-jaco-assistive-robotic-arm}{Jaco}} is a popular sleek 7-DoF robot produced by Kinova Robotics and intended for human assistive applications. As such, it is relatively weak in terms of max torque capabilities. Jaco comes equipped with the \class{JacoThreeFingerGripper} by default, a three-pronged gripper with multi-jointed fingers.}{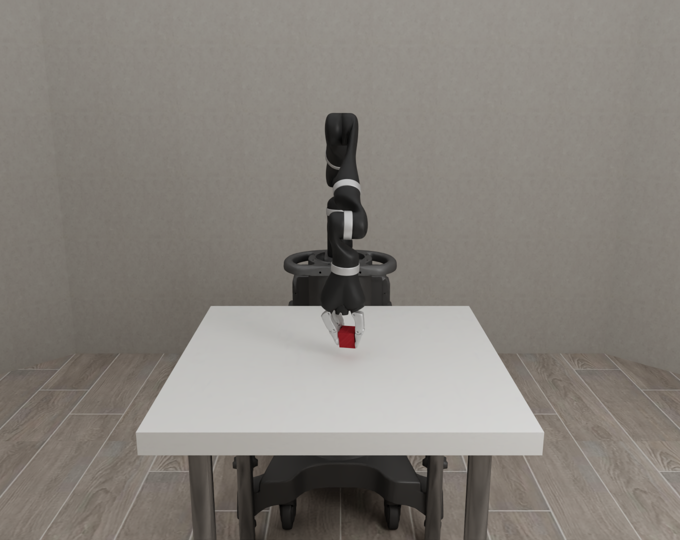}

\robotfigure{\textbf{\href{https://www.kinovarobotics.com/en/products/gen3-robot}{Kinova3}} is Kinova's newest 7-DoF robot, with integrated sensor modules and interfaces designed for research-oriented applications. It is marginally stronger than its Jaco counterpart, and is equipped with the \class{Robotiq85Gripper}, \href{https://robotiq.com/products/2f85-140-adaptive-robot-gripper}{\textbf{Robotiq's 85mm variation}} of their multi-purpose two finger gripper models.}{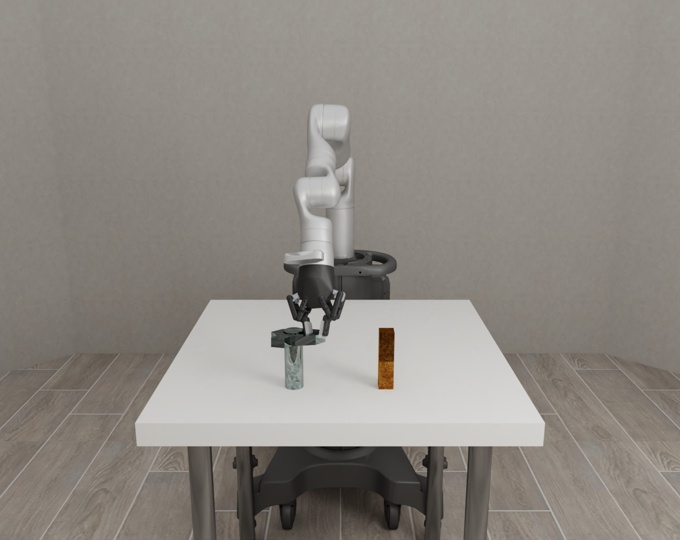}

\robotfigure{\textbf{\href{https://www.universal-robots.com/products/ur5-robot/}{UR5e}} is Universal Robot's newest update to the UR5 line, and is a 6-DoF robot intended for collaborative applications. This newest model boasts an improved footprint and embedded force-torque sensor in its end effector. This arm also uses the \class{Robotiq85Gripper} by default in \projectName{}.}{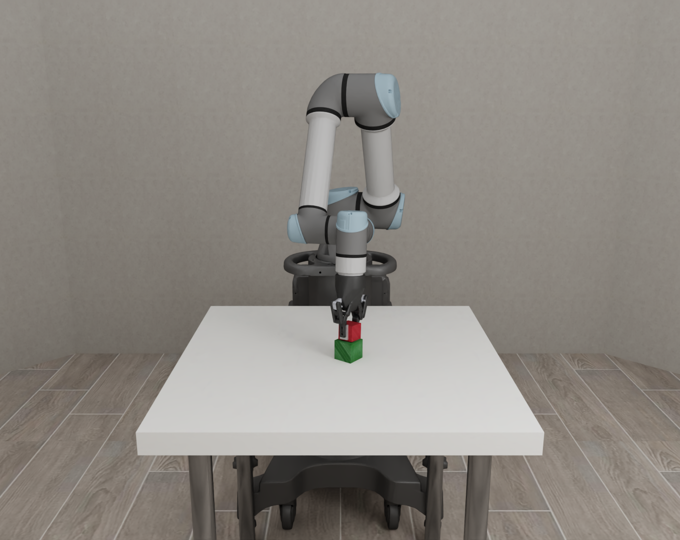}

\robotfigure{\textbf{\href{http://collabrobots.com/about-baxter-robot/}{Baxter}} is an older but classic bimanual robot originally produced by Rethink Robotics but now owned by CoThink Robotics, and is equipped with two 7-DoF arms as well as an addition joint for controlling its swiveling display screen (inactive and disabled by default in \projectName{}). Each arm can be controlled independently in, and is the single multi-armed model currently supported in \projectName{}. Each arm is equipped with a \class{RethinkGripper} by default.}{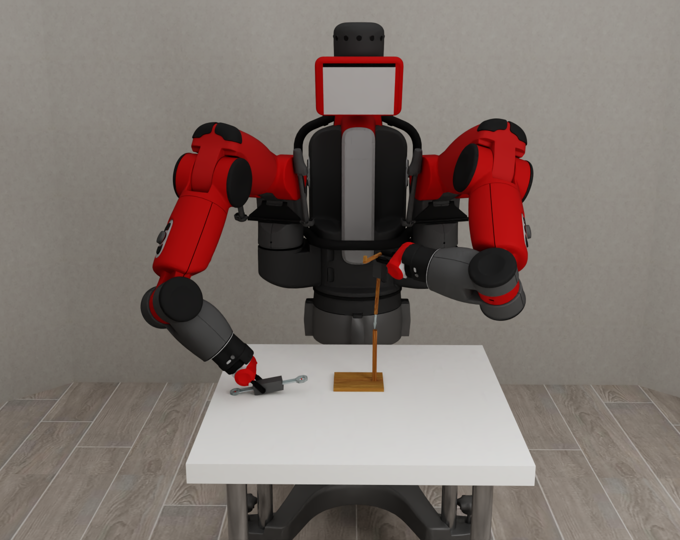}

\robotfigure{\textbf{\href{https://www.fftai.com/products-gr1}{GR1}} is a 44-DoF humanoid robot produced by Fourier Intelligence. Standing 165 cm tall and weighing 55 kg, GR1 has the capability for locomotion, bimanual manipulation, and head movement. GR1 also features cutting-edge vision and sound sensors, enabling intuitive human-like interactions. Attached to each arm is a dexterous hand by default in \projectName{}.}{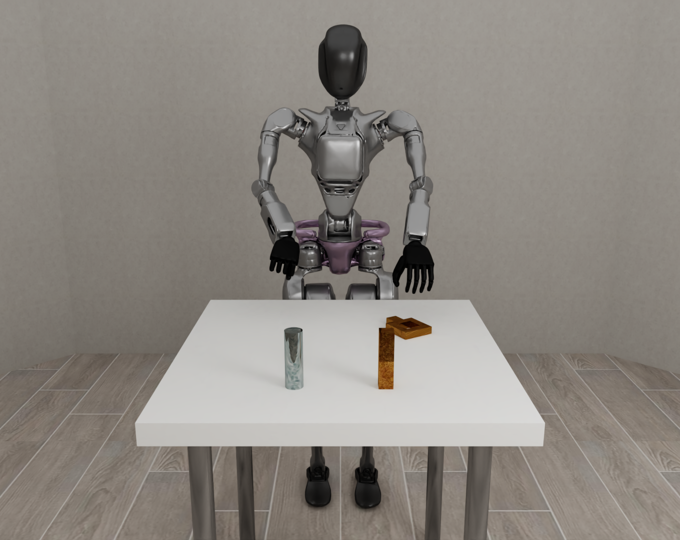}

\robotfigure{\textbf{\href{https://bostondynamics.com/products/spot/}{Spot}} is a 12-DoF, agile, four-legged robot developed by Boston Dynamics. It is capable of navigating complex terrain, avoiding obstacles, and carrying up to 14 kg of payload capacity. In \projectName{}, the Spot robot is equipped with a 6-DoF \href{https://bostondynamics.com/products/spot/arm/}{arm} by default.}{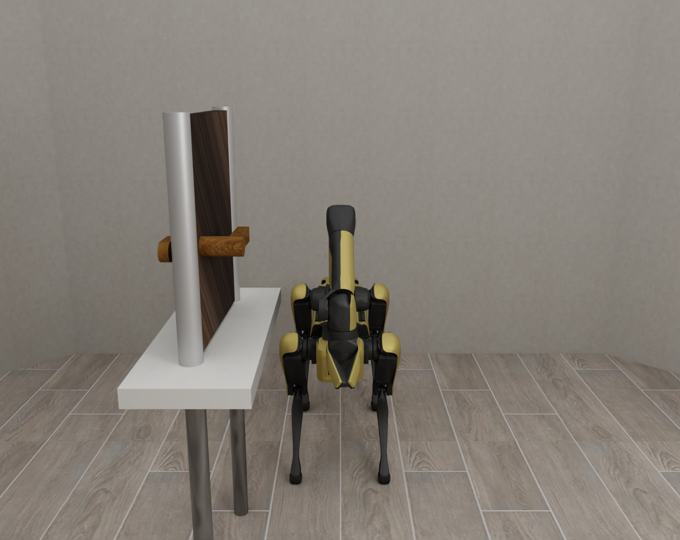}

\robotfigure{\textbf{\href{https://pal-robotics.com/robot/tiago/}{TIAGo}} is a bimanual mobile manipulator robot developed by PAL Robotics. Key features include navigation, interactibility, and a modular design that allows for customization of end effectors, base drives, and computing capabilities. In \projectName{}, the TIAGo robot is equipped with two 7-DoF arms and has mobile base control.}{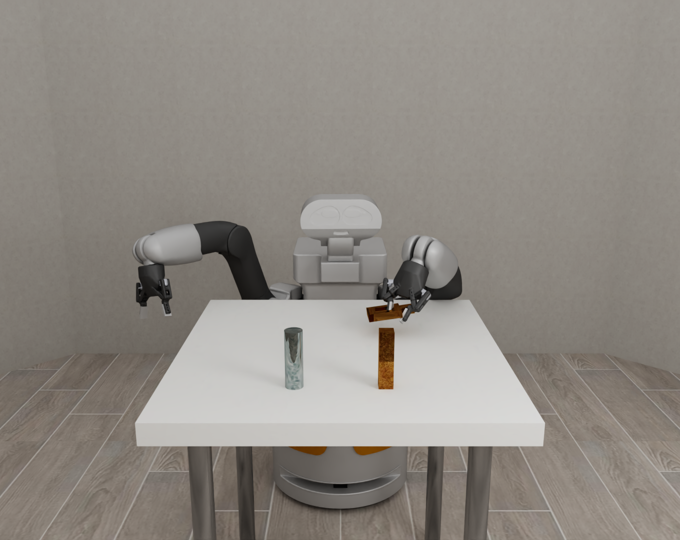}

\subsection{Controllers}
\label{sec:controllers}

\textbf{Controllers} transform the high-level actions into low-level virtual motor commands that actuate the robots. The high-level actions are interpreted as reference signals for the controllers, i.e., desired configurations to be reached. Underlying all our robot models is a set of virtual motors actuated at each joint that execute given torques. The controllers will translate the reference signals into corresponding joint torque values to try to achieve that desired configuration. Starting from \texttt{robosuite} v1.5, we employ \textbf{composite} controllers. A composite controller takes in a high-level action vector and converts it into commands for each \textbf{body part} controller, where each body part's controller is responsible for translating supplied actions into joint torques for that specific part. This design enables modularity when controlling robots that can be decomposed into multiple body parts. For example, the Operational Space Controllers~\cite{khatib1995inertial} interpret high-level actions as desired end-effector configurations, either positions (three degrees of freedom) or poses (six degrees of freedom), and compute the corresponding joint torques to move the robot's end-effector from its current pose to a desired one. 
Our controllers facilitate sim-to-real transferability, as torque-based controllers are common to most real-world existing robotic platforms such as Rethink Robotics Sawyer, Franka Panda, Kuka IIWA, and Kinova Jaco, and enables contact-rich manipulation with control of the interaction forces.

\begin{table}[]
\begin{adjustbox}{angle=90}
\footnotesize
\begin{tabular}{|c|c|c|c|}
\cline{1-4}
\begin{tabular}{c}Controller Name \\ and Options\end{tabular} & Controller Type & \begin{tabular}{c}Action Dimensions \\ (Gripper Not Included)\end{tabular} & Action Format\\
\cline{1-4}
\begin{tabular}{c}OSC\_POSE \\ impedance\_mode $=$ fixed\end{tabular} & \begin{tabular}{c}Operational Space Control \\ (Position \& Orientation)\end{tabular} & 6 & $(p_x, p_y, p_z, r_x, r_y, r_z)$ \\
\cline{1-4}
\begin{tabular}{c}OSC\_POSE \\ impedance\_mode $=$ variable\_kp\end{tabular} & \begin{tabular}{c}Operational Space Control \\ with variable stiffness \\ (critically damped)\end{tabular} & 12 & \begin{tabular}{c}$(p_x, p_y, p_z, r_x, r_y, r_z)$\\$(k^p_{px},k^p_{py},k^p_{pz},k^p_{rx},k^p_{ry},k^p_{rz})$ \end{tabular}\\
\cline{1-4}
\begin{tabular}{c}OSC\_POSE \\ impedance\_mode $=$ variable\end{tabular} & \begin{tabular}{c}Operational Space Control \\ with variable impedance\end{tabular} & 18 & \begin{tabular}{c}$(p_x, p_y, p_z, r_x, r_y, r_z)$\\$(k^p_{px},k^p_{py},k^p_{pz},k^p_{rx},k^p_{ry},k^p_{rz})$ \\$(k^d_{px},k^d_{py},k^d_{pz},k^d_{rx},k^d_{ry},k^d_{rz})$\end{tabular}\\
\cline{1-4}
\begin{tabular}{c}OSC\_POSITION \\ impedance\_mode $=$ fixed\end{tabular} & \begin{tabular}{c}Operational Space Control \\ (Position only)\end{tabular} & 3 & $(p_x, p_y, p_z)$ \\
\cline{1-4}
\begin{tabular}{c}OSC\_POSITION \\ impedance\_mode $=$ variable\_kp\end{tabular} & \begin{tabular}{c}Operational Space Control \\ with variable stiffness \\ (critically damped)\end{tabular} & 9 & \begin{tabular}{c}$(p_x, p_y, p_z)$\\$(k^p_{px},k^p_{py},k^p_{pz},k^p_{rx},k^p_{ry},k^p_{rz})$ \end{tabular}\\
\cline{1-4}
\begin{tabular}{c}OSC\_POSITION \\ impedance\_mode $=$ variable\end{tabular} & \begin{tabular}{c}Operational Space Control \\ with variable impedance\end{tabular} & 15 & \begin{tabular}{c}$(p_x, p_y, p_z)$\\$(k^p_{px},k^p_{py},k^p_{pz},k^p_{rx},k^p_{ry},k^p_{rz})$ \\$(k^d_{px},k^d_{py},k^d_{pz},k^d_{rx},k^d_{ry},k^d_{rz})$\end{tabular}\\
\cline{1-4}
IK\_POSE & \begin{tabular}{c}Inverse Kinematics Control \\ (Position \& Orientation)\end{tabular} & 7 & $(p_x, p_y, p_z, q_x, q_y, q_z, q_w)$ \\
\cline{1-4}
\begin{tabular}{c}JOINT\_POSITION \\ impedance\_mode $=$ fixed\end{tabular} & Joint Position Control & n & n joint positions \\
\cline{1-4}
\begin{tabular}{c}JOINT\_POSITION \\ impedance\_mode $=$ variable\_kp\end{tabular} & \begin{tabular}{c}Joint Position Control \\ with variable stiffness \\ (critically damped)\end{tabular} & 2n & \begin{tabular}{c}n joint positions\\ and $k^p$ for each joint \end{tabular}\\
\cline{1-4}
\begin{tabular}{c}JOINT\_POSITION \\ impedance\_mode $=$ variable\end{tabular} & \begin{tabular}{c}Joint Position Control\\ with variable impedance\end{tabular} & 3n & \begin{tabular}{c}n joint positions\\and $(k^p,k^d)$ for each joint\end{tabular}\\
\cline{1-4}
JOINT\_VELOCITY& Joint Velocity Control & n & n joint velocities\\
\cline{1-4}
JOINT\_TORQUE& Joint Torque Control & n & n joint torques\\
\cline{1-4}
\end{tabular}
\end{adjustbox}
\caption{Body Part Controller Configurations available in \projectName{}}
\label{tab:controllers}
\end{table}

We include the following composite controller options as part of \projectName{}: \texttt{BASIC} and \texttt{WHOLE\_BODY\_IK}. The \texttt{BASIC} composite controller directly splits and passes down the high level action vector to the individual body part controllers that operate independently to control various parts of the robot, such as arms, torso, head, base, and legs. Each part can be assigned a specific body part controller type (e.g., \texttt{OSC\_POSE} and \texttt{JOINT\_POSITION}) depending on the desired control behavior for that part. For example, arms may use \texttt{OSC\_POSE} for precise end-effector control, while the base may use \texttt{JOINT\_VELOCITY} for movement across the ground. The \texttt{WHOLE\_BODY\_IK} composite controller assumes that the high level action vector contains end effector pose targets and uses an inverse kinematics solver to compute joint angles to reach those pose targets. Then, the composite controller passes the joint angles down to the corresponding \texttt{JOINT\_POSITION} body part controllers. Finally, users can leverage custom third-party composite controllers, by subclassing the \class{CompositeController} class and implementing the custom methods. We provide an example of a third-party composite controller implementation, \texttt{WHOLE\_BODY\_MINK\_IK} \cite{Zakka_Mink_Python_inverse_2024}, in \projectName{}.

We include the following body part controller options as part of \projectName{}: \texttt{OSC\_POSE}, \mbox{\texttt{OSC\_POSITION}}, \texttt{JOINT\_POSITION}, \texttt{JOINT\_VELOCITY}, and \texttt{JOINT\_TORQUE} (see Table~\ref{tab:controllers}). For \texttt{OSC\_POSE}, \mbox{\texttt{OSC\_POSITION}}, and \texttt{JOINT\_POSITION}, we include three variants: First, the most common variant is to use a predefined and \textbf{fixed} set of impedance parameters (\texttt{impedance\_mode} $=$ \texttt{fixed}). In this case, the action space only includes the desired pose, position, or joint configuration. The second options is to control the \textbf{stiffness} of the actuation (\texttt{impedance\_mode} $=$ \texttt{variable\_kp}), i.e., with how much force will the robot react to deviations to the desired configuration. This is controlled via the proportional parameters of the controller ($k_p$). The damping parameters ($k_d$) are automatically set to the values that lead to a critically damped system. The third variant allows full control of the \textbf{impedance} behavior (\texttt{impedance\_mode} $=$ \texttt{variable}), with the actions including both stiffness and damping parameters. These last two variants lead to extended action spaces that can control the stiffness and damping behavior of the controller in a variable manner over the course of an interaction, providing full control to the policy/solution over the contact and dynamic behavior of the robot.

For the \texttt{OSC\_POSITION} variants, the robot will hold the initial orientation while trying to achieve the position given in the action. Variants controlling stiffness, or stiffness and damping can specify not only these parameters for the position but also for orientation. Therefore, the dimensionality of the action spaces with these controllers are 9 and 15 (row 6 and 7 in Table~\ref{tab:controllers}). 

\subsection{Objects}
\label{sec:objects}

\textbf{Objects}, such as boxes and cans, are essential to building manipulation environments. We designed the \class{MujocoObject} interfaces to standardize and simplify the procedure for importing 3D models into the scene or procedurally generate new objects. MuJoCo defines models via the MJCF XML format. These MJCF files can either be stored as XML files on disk and loaded into simulator, called \class{MujocoXMLObject}, or be created on-the-fly by code prior to simulation, called \class{MujocoGeneratedObject}. Based on these two mechanisms of how MJCF models are created, we offer two main ways of creating your own object:
\begin{itemize}
    \item Define an object in an MJCF XML file;
    \item Use procedural generation APIs to dynamically create an MJCF model.
\end{itemize}

In the former case, an object can be defined using MuJoCo's native \href{http://www.mujoco.org/book/XMLreference.html}{\textbf{MJCF format}} and can be loaded directly into the simulation using \projectName{}'s API. In the latter case, a complex object can be defined by sequentially composing a set of primitive geoms and defining their poses relative to the rest of the object. Additionally, \projectName{} supports custom texture definitions to be added to specific geoms defined. We refer the interested reader to the \class{HammerObject} class as a showcase example for procedurally-generated objects.

\subsection{Sensors}
\label{sec:sensors}

The simulator generates virtual physical signals as response to a robot’s interactions through \textbf{Sensors}. Virtual signals include images, force-torque measurements (from a force-torque sensor like the one included by default in the wrist of all \class{Gripper} models), pressure signals (e.g., from a sensor on the robot’s finger or on the environment), etc. Sensors, except cameras and joint sensors, are accessed via the function \texttt{get\_sensor\_measurement}, by providing the name of the sensor.

Every robot joint provides information about its state, including position and velocity. In MuJoCo these are not measured by sensors, but resolved and set by the simulator as the result of the actuation forces. Therefore, they are not accessed through the common \texttt{get\_sensor\_measurement} function but as properties of the \class{Robot} simulation API, for instance, \texttt{\_joint\_positions} and \texttt{\_joint\_velocities}.

Cameras bundle a name to a set of properties to render images of the environment such as the pose and pointing direction, field of view, and resolution. Inheriting from MuJoCo, cameras are defined in the \class{RobotModel} and \class{Arena} models and can be attached to any body. Images, as they would be generated from the cameras, are not accessed through \texttt{get\_sensor\_measurement} but via the renderer (e.g., OpenGL-based \href{https://mujoco.readthedocs.io/en/stable/programming/visualization.html#opengl-rendering}{\textbf{MjViewer}} or \href{https://www.pygame.org/news}{\textbf{PyGame}}). In a common user pipeline, images are not queried directly; we specify one or several cameras we want to use images from when we create the environment, and the images are generated and appended automatically to the observation dictionary.

\subsection{I/O Devices}
\label{sec:devices}

\textbf{Devices} define the external controllers a user can teleoperate robots in real-time. A \class{Device} object reads user input from I/O devices, such as a \class{Keyboard}, \class{SpaceMouse} or \class{MJGUI}, and parse it into control commands sent to the robots. The \class{MJGUI} device uses MuJoCo's built in GUI and users' cursor to drag and drop certain pre-defined robot controller targets. The \href{https://3dconnexion.com/uk/spacemouse/}{\textbf{SpaceMouse}} from 3Dconnexion is a 3D mouse that we used extensively for collecting demonstrations~\cite{zhu2018reinforcement, mandlekar2018roboturk} and debugging task designs. More generally, we support any interface that implements the \class{Device} abstract class. To support your own custom device, simply subclass this base class and implement the required methods.

\section{Benchmark Environments}

\newcommand{\envfigure}[2]{
\begin{figure}[H]
\floatbox[{\capbeside\thisfloatsetup{capbesideposition={right,top},
capbesidewidth=0.7\textwidth}}]{figure}[\FBwidth]
{\caption*{#1}}
{\includegraphics[width=0.3\textwidth]{#2}}
\end{figure}
}

\newcommand{\benchfigure}[2]{
\begin{figure}[H]
    \centering
    \includegraphics[width=\linewidth]{#2}
    \caption{#1}
\end{figure}
}

\subsection{Task Descriptions}
\label{sec:task_desc}

We provide a brief description of each environment below, along with a sequence of frames that depict a successful rollout.


\envfigure{
\textbf{Block Lifting:} A cube is placed on the tabletop in front of a single robot arm. The robot arm must lift the cube above a certain height. The cube location is randomized at the beginning of each episode.
}{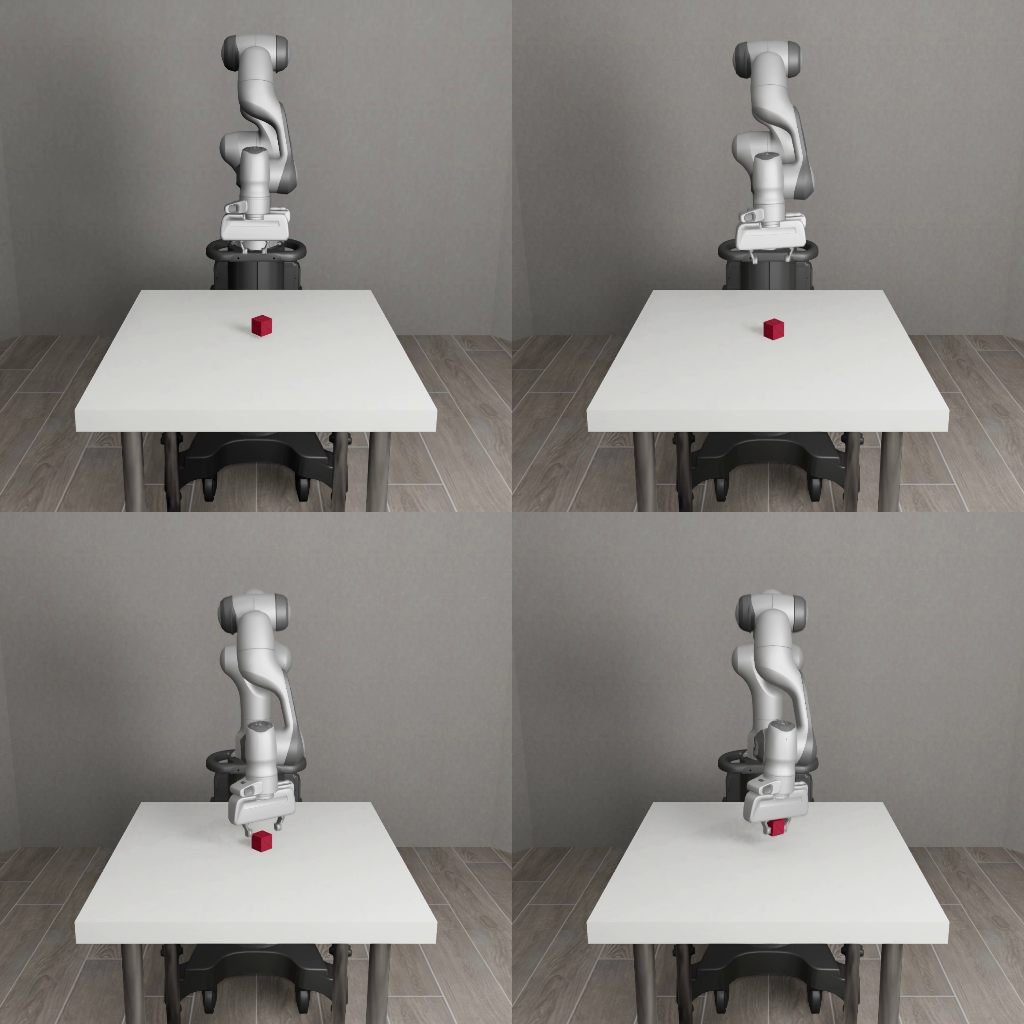}

\envfigure{
\textbf{Block Stacking:} Two cubes are placed on the tabletop in front of a single robot arm. The robot must place one cube on top of the other cube. The cube locations are randomized at the beginning of each episode.
}{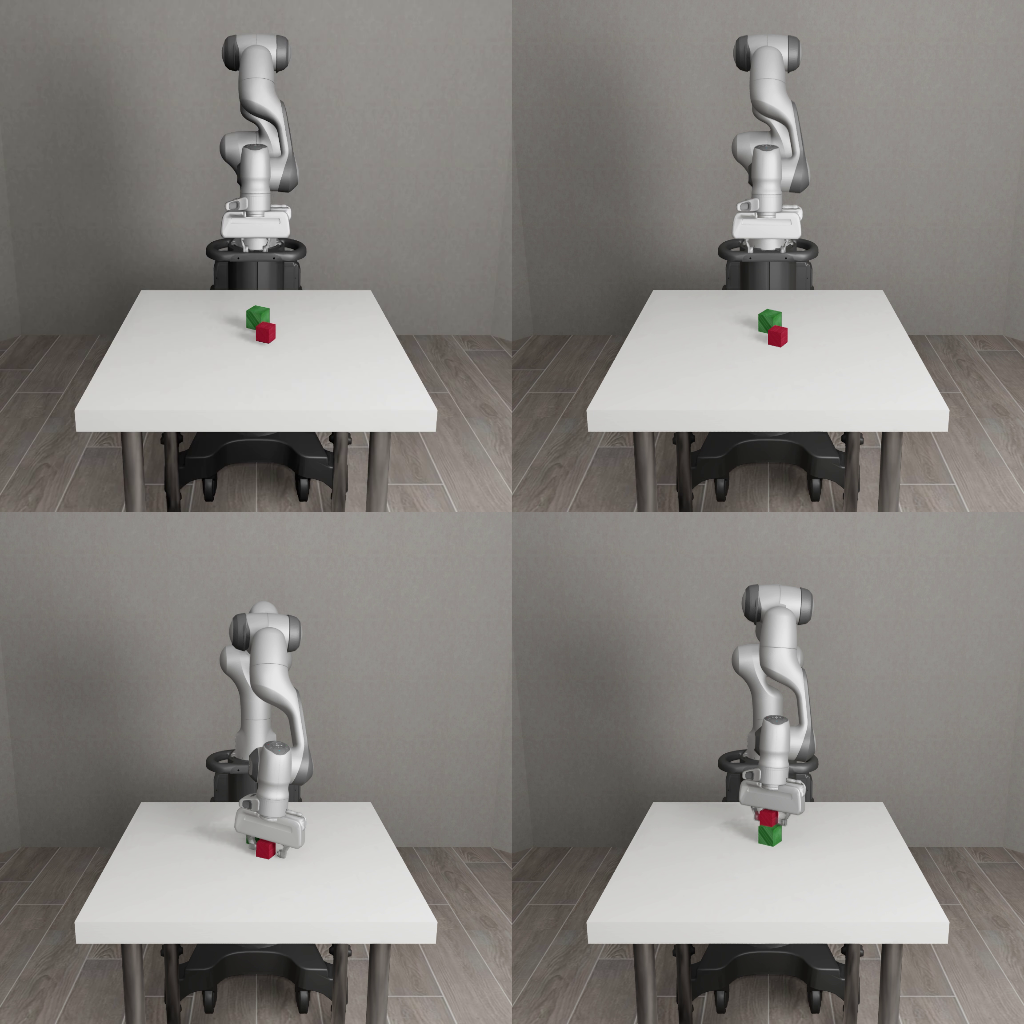}

\envfigure{
\textbf{Pick-and-Place:} Four objects are placed in a bin in front of a single robot arm. There are four containers next to the bin. The robot must place each object into its corresponding container. This task also has easier single-object variants. The object locations are randomized at the beginning of each episode.
}{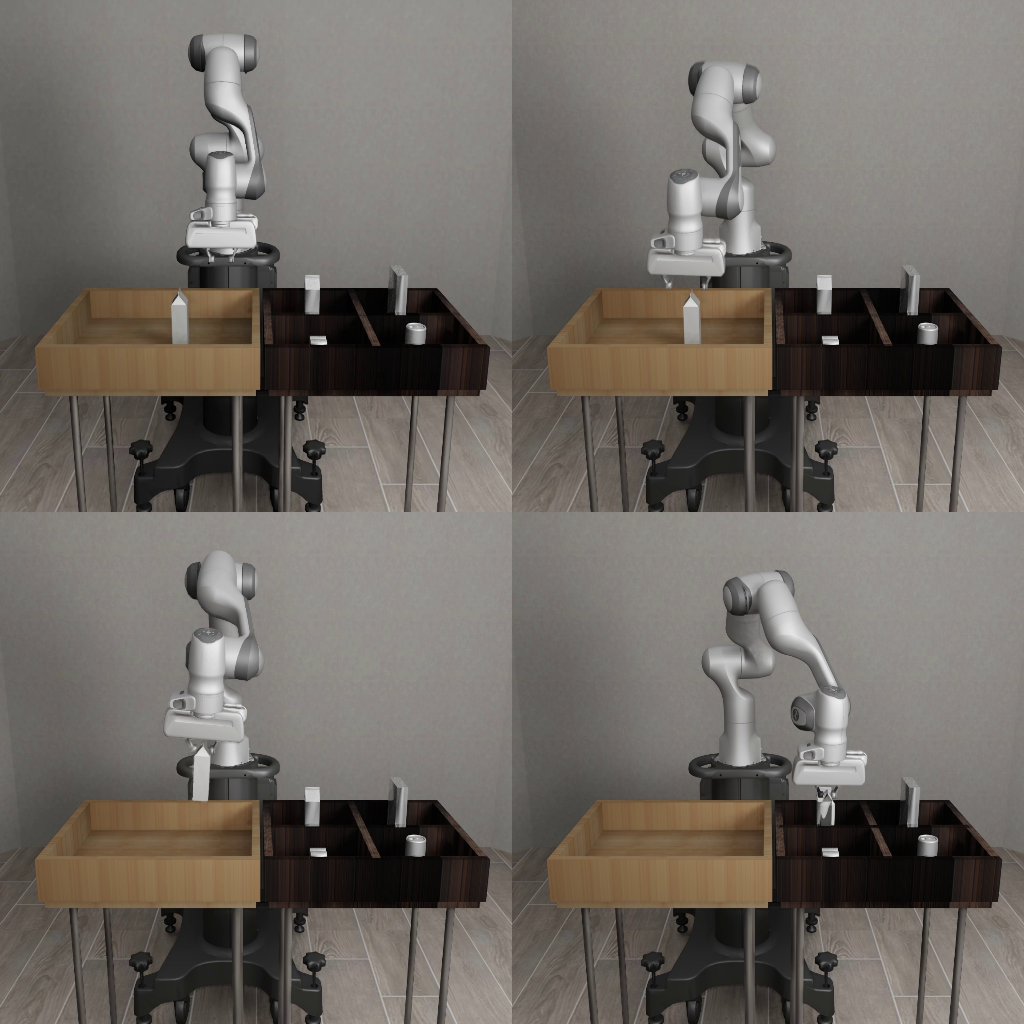}

\envfigure{
\textbf{Nut Assembly:} Two colored pegs (one square and one round) are mounted on the tabletop, and two colored nuts (one square and one round) are placed on the table in front of a single robot arm. The robot must fit the square nut onto the square peg and the round nut onto the round peg. This task also has easier single nut-and-peg variants. The nut locations are randomized at the beginning of each episode.
}{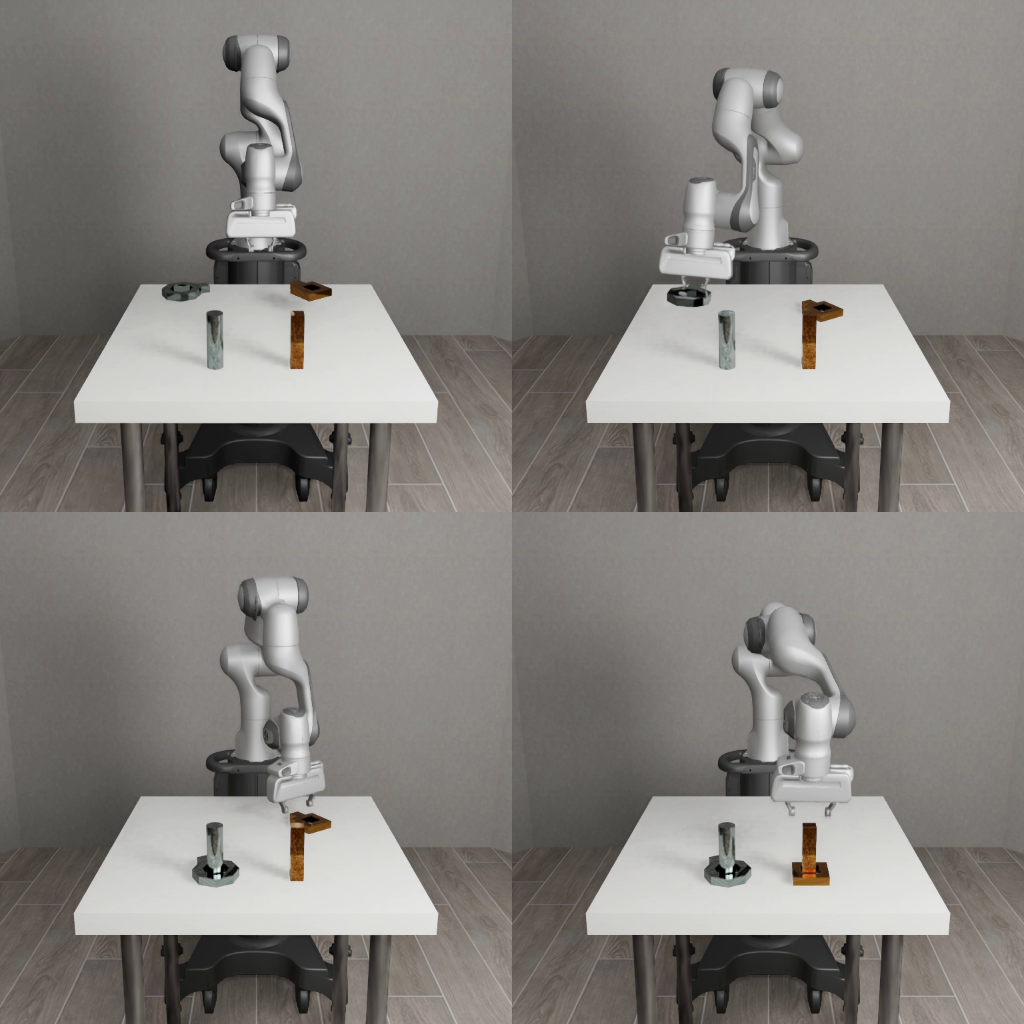}

\envfigure{
\textbf{Door Opening:} A door with a handle is mounted in free space in front of a single robot arm. The robot arm must learn to turn the handle and open the door. The door location is randomized at the beginning of each episode.
}{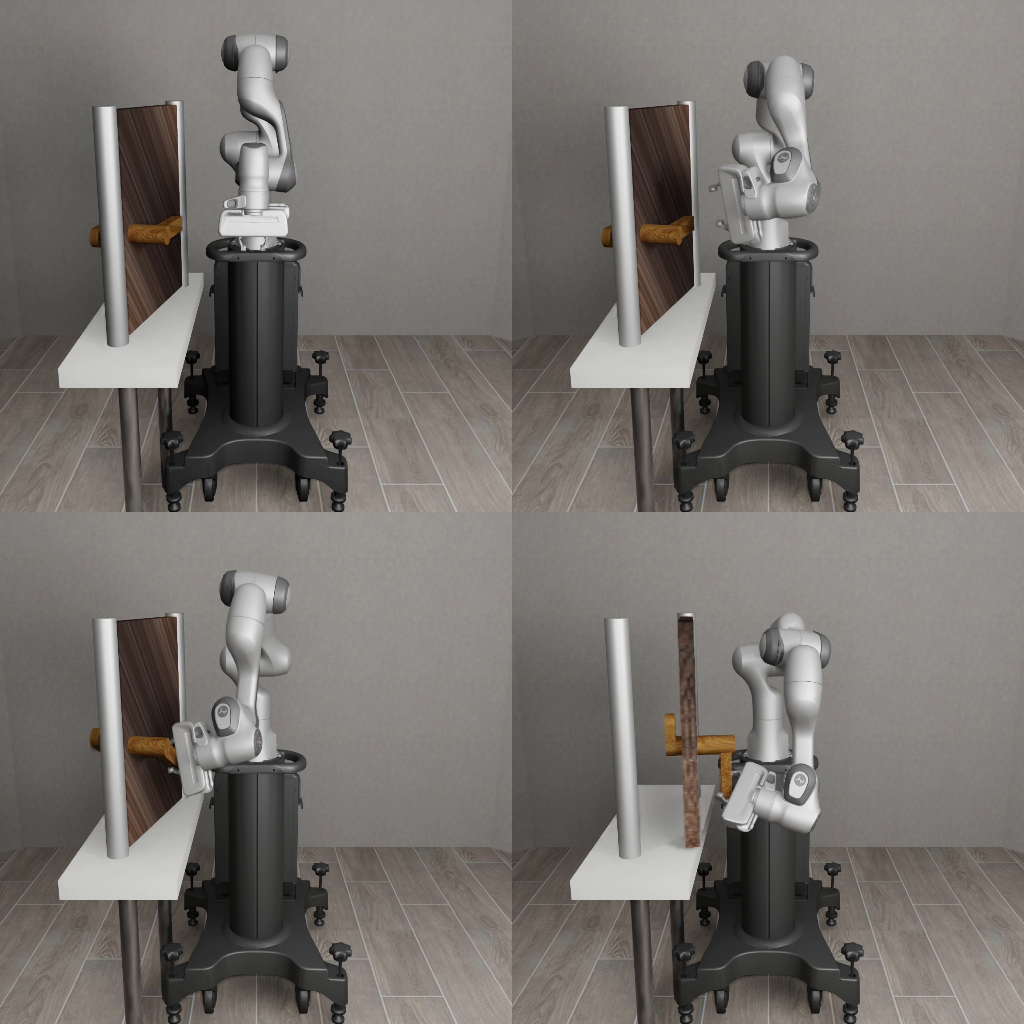}

\envfigure{
\textbf{Table Wiping:} A table with a whiteboard surface and some markings is placed in front of a single robot arm, which has a whiteboard eraser mounted on its hand. The robot arm must learn to wipe the whiteboard surface and clean all of the markings. The whiteboard markings are randomized at the beginning of each episode.
}{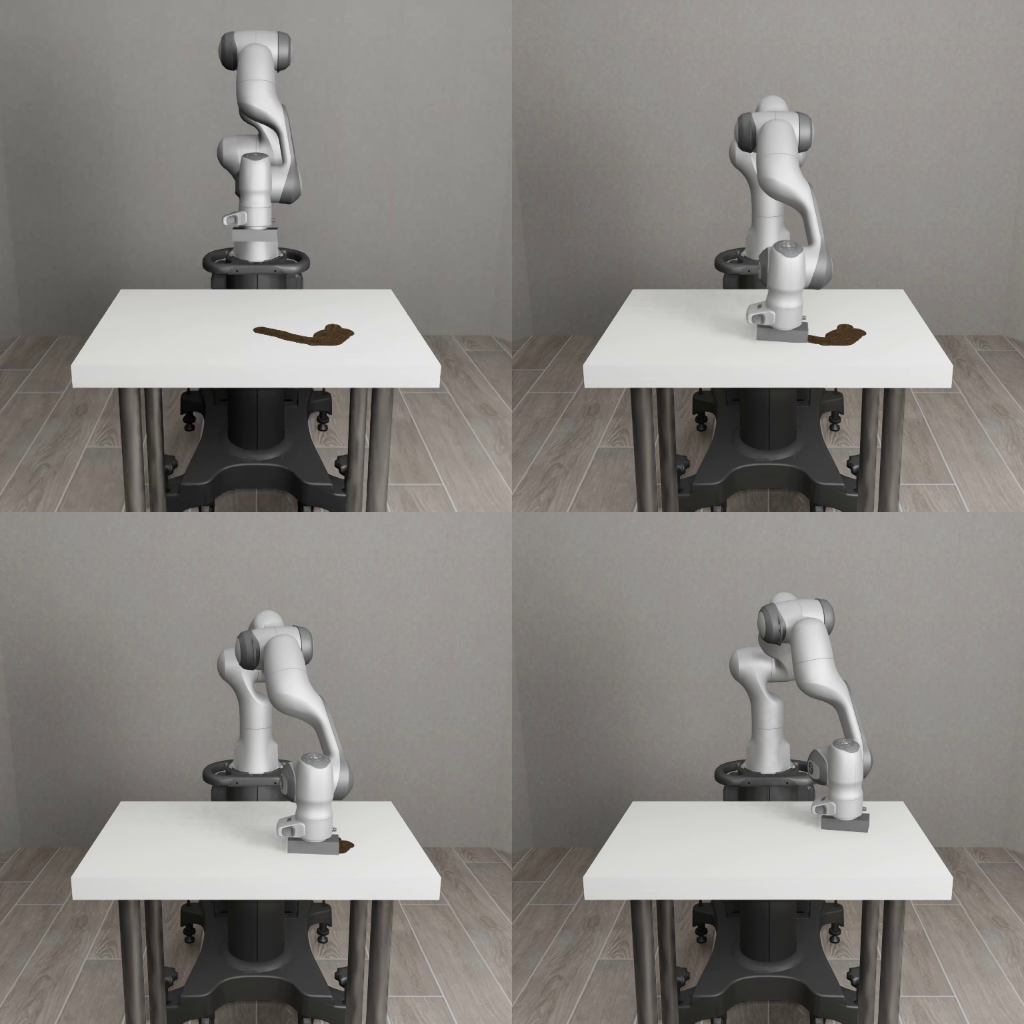}

\envfigure{
\textbf{Two Arm Lifting:} A large pot with two handles is placed on a table top. Two robot arms are placed on the same side of the table or on opposite ends of the table. The two robot arms must each grab a handle and lift the pot together, above a certain height, while keeping the pot level. The pot location is randomized at the beginning of each episode.
}{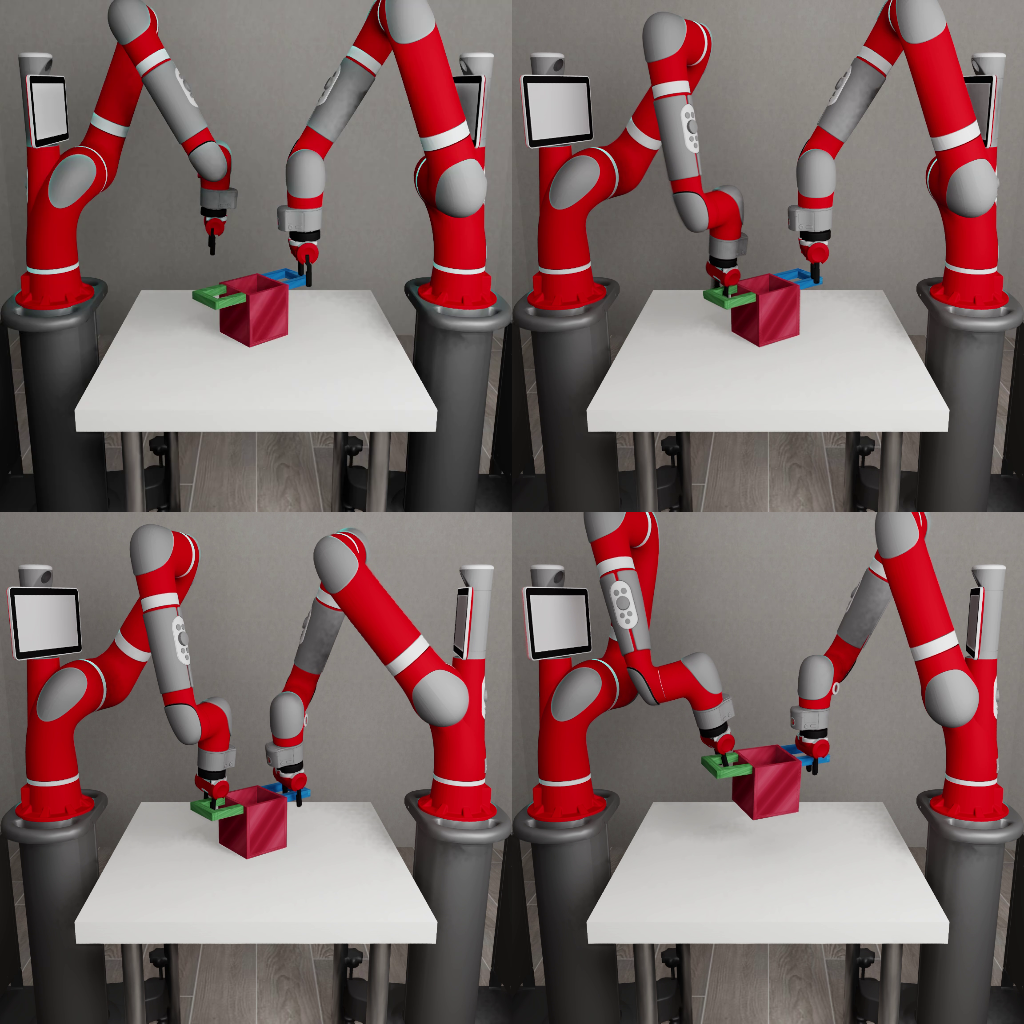}

\envfigure{
\textbf{Two Arm Peg-In-Hole:} Two robot arms are placed either next to each other or opposite each other. One robot arm holds a board with a square hole in the center, and the other robot arm holds a long peg. The two robot arms must coordinate to insert the peg into the hole. The initial arm configurations are randomized at the beginning of each episode.
}{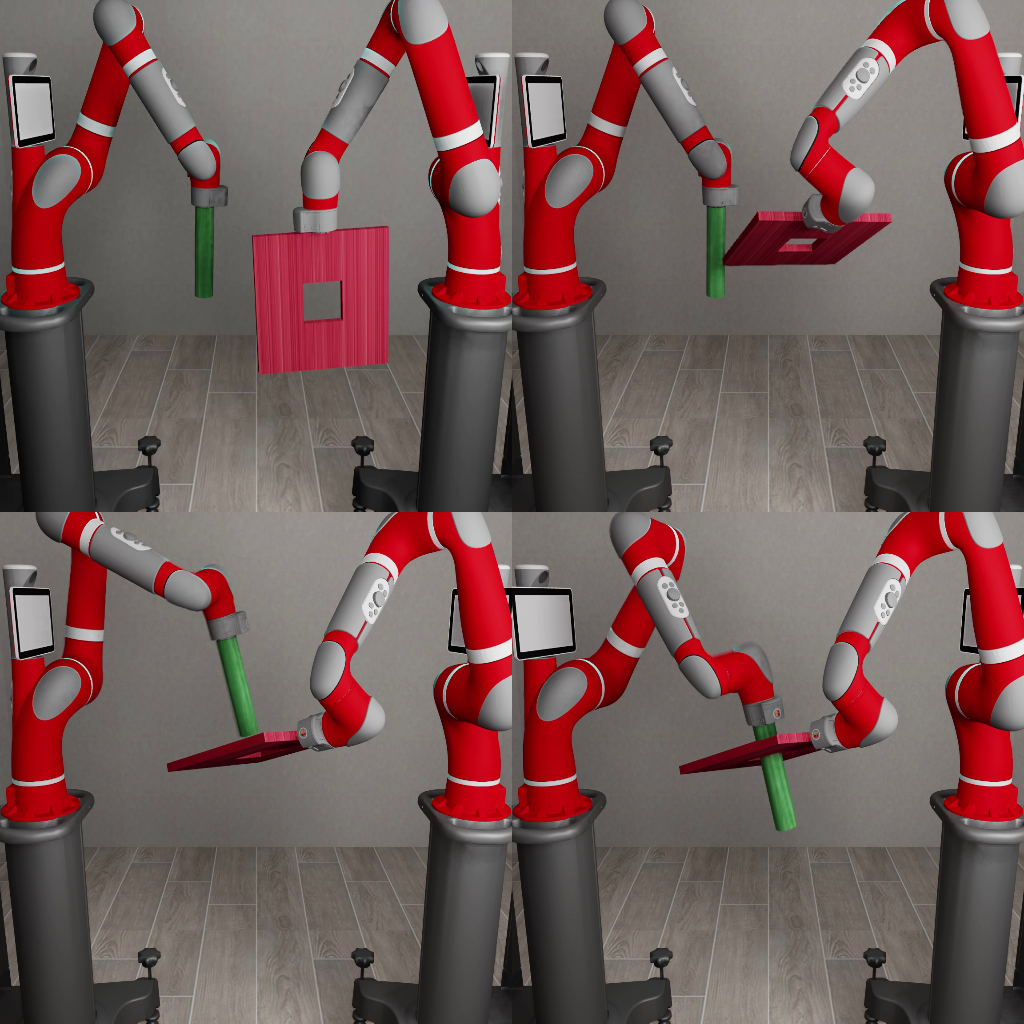}

\envfigure{
\textbf{Two Arm Handover:} A hammer is placed on a narrow table. Two robot arms are placed on the same side of the table or on opposite ends of the table. The two robot arms must coordinate so that the arm closer to the hammer picks it up and hands it to the other arm. The hammer location and size is randomized at the beginning of each episode.
}{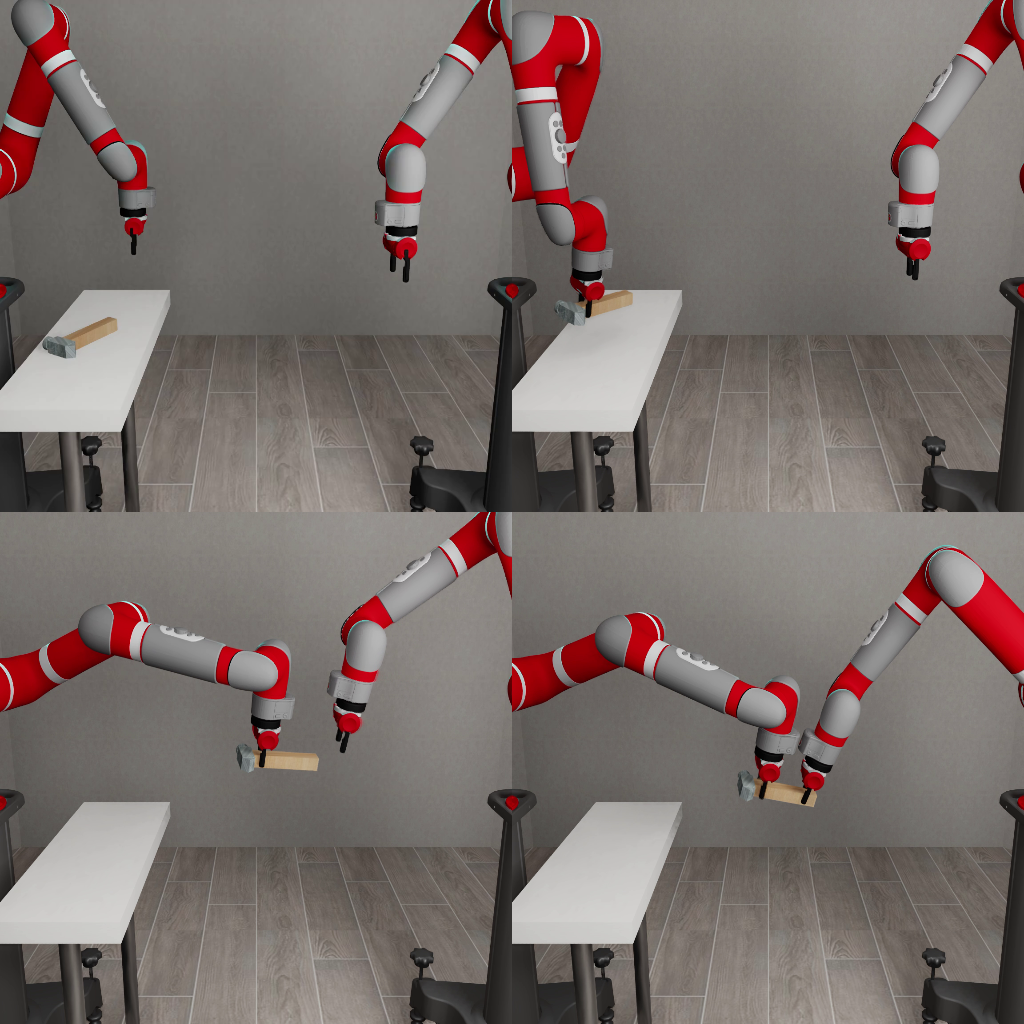}

\subsection{Benchmarking Results}
\label{sec:benchmarking_results}

We provide a standardized set of benchmarking experiments as baselines for future experiments. Specifically, we test Soft Actor-Critic (SAC)~\cite{haarnoja2018soft}, the state-of-the-art model-free RL algorithm, on a select combination of tasks (all) using a combination of proprioceptive and object-specific observations, robots (\class{Panda}, \class{Sawyer}), and controllers (\texttt{OSC\_POSE}, \texttt{JOINT\_VELOCITY}). Our experiments were implemented and executed in an extended version of \href{https://github.com/vitchyr/rlkit}{\textbf{rlkit}}, a popular PyTorch-based RL framework and algorithm library. For ease of replicability, we have released our official experimental results on a \href{https://github.com/ARISE-Initiative/robosuite-benchmark}{\textbf{benchmark repository}}.

\begin{figure}
    \centering
    
    \begin{subfigure}[b]{0.32\textwidth}
         \centering
         \includegraphics[width=1.0\textwidth]{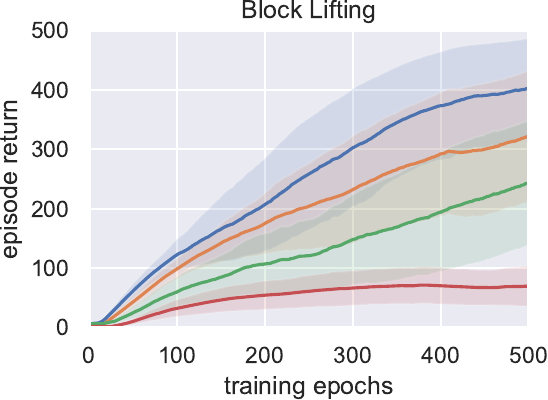}
     \end{subfigure}
     \begin{subfigure}[b]{0.32\textwidth}
         \centering
         \includegraphics[width=1.0\textwidth]{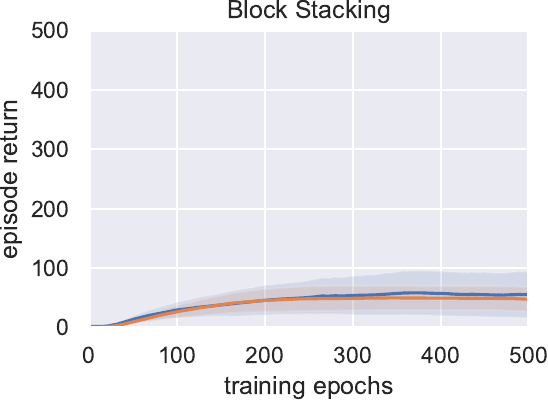}
     \end{subfigure}
     \begin{subfigure}[b]{0.32\textwidth}
         \centering
         \includegraphics[width=1.0\textwidth]{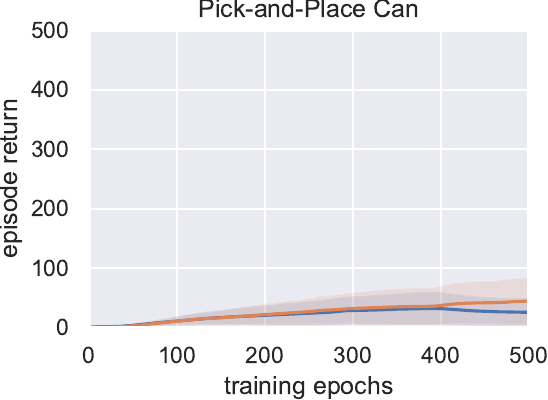}
     \end{subfigure}
     \vspace{3mm}
     
     \begin{subfigure}[b]{0.32\textwidth}
         \centering
         \includegraphics[width=1.0\textwidth]{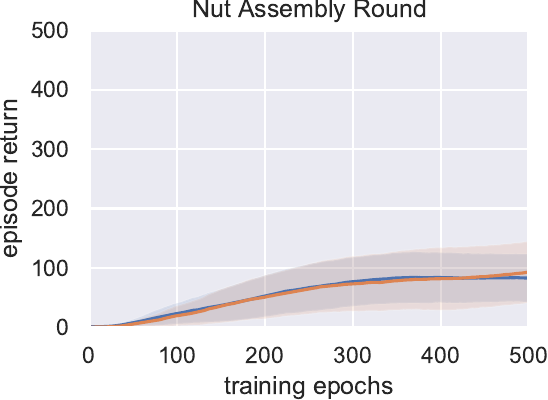}
     \end{subfigure}
     \begin{subfigure}[b]{0.32\textwidth}
         \centering
         \includegraphics[width=1.0\textwidth]{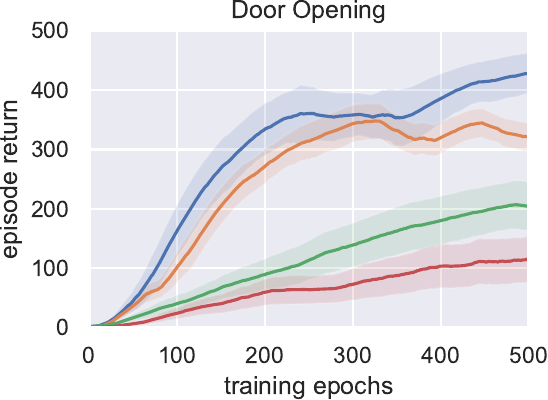}
     \end{subfigure}
     \begin{subfigure}[b]{0.32\textwidth}
         \centering
         \includegraphics[width=1.0\textwidth]{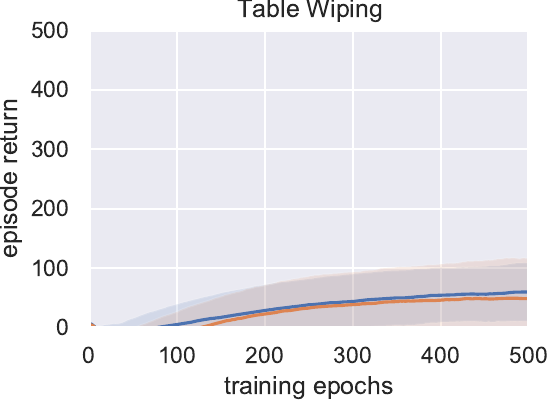}
     \end{subfigure}
     \vspace{3mm}

     \begin{subfigure}[b]{0.32\textwidth}
         \centering
         \includegraphics[width=1.0\textwidth]{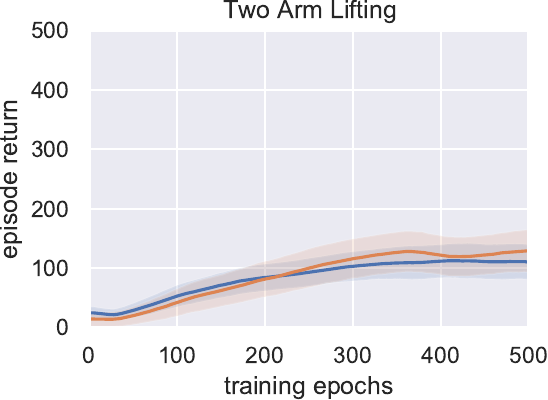}
     \end{subfigure}
     \begin{subfigure}[b]{0.32\textwidth}
         \centering
         \includegraphics[width=1.0\textwidth]{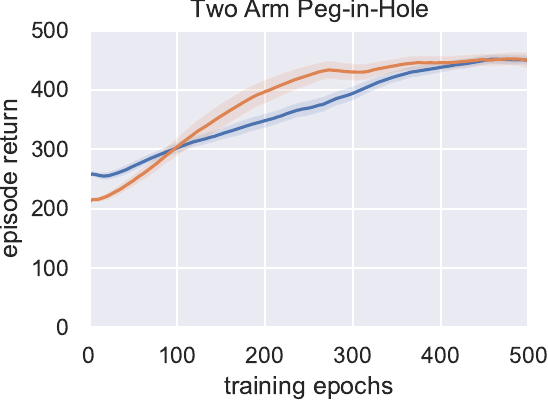}
     \end{subfigure}
     \begin{subfigure}[b]{0.32\textwidth}
         \centering
         \includegraphics[width=1.0\textwidth]{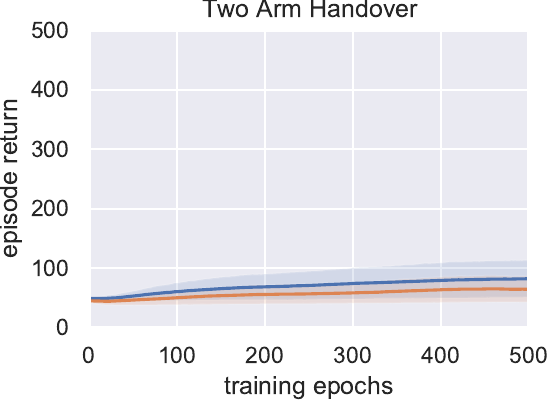}
     \end{subfigure}
     \vspace{2mm}
     
     \begin{subfigure}[b]{0.9\textwidth}
         \centering
         \includegraphics[width=1.0\textwidth]{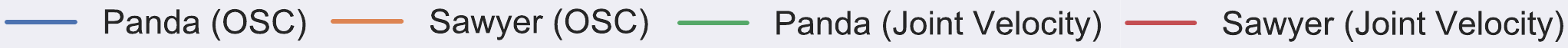}
     \end{subfigure}

    \caption{Benchmarking results on the nine standardized environments in \projectName{}. For the Two Arm tasks, we use two Panda arms for Panda (OSC) and two Sawyer arms for Sawyer (OSC)}
    \label{fig:benchmarking}
\end{figure}

All agents were trained for 500 epochs with 500 steps per episode, and utilize the same standardized algorithm hyperparameters (see our benchmarking repo above for exact parameter values). The agents receive the low-dimensional physical states as input to the policy. These experiments ran on 2 CPUs and 12G VRAM and no GPU, each taking about two days to complete. We normalize the per-step rewards to 1.0 such that the maximum possible per-episode return is 500. In Figure~\ref{fig:benchmarking}, we show the per-task experiments conducted, with each experiment's training curve showing the evaluation return mean's average and standard deviation over five random seeds. 

We select two of the easiest environments, \textbf{Block Lifting} and \textbf{Door Opening}, for an ablation study between the operational space controllers (\texttt{OSC\_POSE}) and the joint velocity controllers (\texttt{JOINT\_VELOCITY}). We observe that the choice of controllers alone has an evident impact on the efficiency of learning. Both robots learn to solve the tasks faster with the operational space controllers, which we hypothesize is credited to the accelerated exploration in task space; this highlights the potential of this impedance-based controller to improve task performance on robotic tasks that were previously limited by their action space parameterization. The SAC algorithm is able to solve three of the nine environments, including \textbf{Block Lifting}, \textbf{Door Opening}, and \textbf{Two Arm Peg-in-Hole}, while making slow progress in the other environments, which requires intelligent exploration in longer task horizons. For future experiments, we recommend using the nine environments with the Panda robot and the operational space controller, i.e., the blue curves of Panda (OSC) in Figure~\ref{fig:benchmarking}, for standardized and fair comparisons.


\section{Conclusion}

\projectName{} provides a simulation framework and benchmark of environments for research and development of robot learning solutions. It includes a suite of standardized tasks for rigorous evaluation and reproducible research. This framework is built on top of the \href{http://www.mujoco.org/}{\textbf{MuJoCo}} physics engine. Since its debut in 2017, robosuite has been used by the AI and robotics research community in a variety of topics, including reinforcement learning~\cite{corl2018surreal, martin2019variable}, imitation learning~\cite{mandlekar2018roboturk}, sim-to-real transfer~\cite{martin2019variable}, etc. With the presented v1.5 version, we hope to facilitate more diverse research and reproducible and benchmarkable advances. We invite the community to benchmark their solutions in the provided tasks, and to contribute to robosuite for future releases.

\subsection*{Acknowledgement}
\label{sec:acknowledgement}
 We would like to thank members of the Stanford People, AI, \& Robots (PAIR) Group for their support and feedback to \projectName{}. In particular, the following people have made their contributions in different stages of this project: Jiren Zhu, Joan Creus-Costa, Jim Fan, Zihua Liu, Orien Zeng, Anchit Gupta, Michelle Lee, Rachel Gardner, Danfei Xu, Andrew Kondrich, Jonathan Booher, Albert Tung, Zhenyu Jiang, Yuqi Xie, You Liang Tan. We wholeheartedly welcome the community to contribute to \projectName{} through issues and pull requests. New contributors will be listed on our project website: \href{https://robosuite.ai}{\textbf{robosuite.ai}}

\bibliographystyle{plain}
\bibliography{main}

\end{document}